\documentclass[11pt]{article}

\usepackage[preprint]{acl2026}

\usepackage{times}
\usepackage{latexsym}
\usepackage{float}

\usepackage[T1]{fontenc}

\usepackage[utf8]{inputenc}

\usepackage{microtype}

\usepackage{inconsolata}

\usepackage[frozencache,cachedir=minted-cache]{minted}
\usepackage{graphicx}
\usepackage{multirow}
\usepackage{multicol}
\usepackage{placeins}
\usepackage{amsmath}
\usepackage{algorithm}
\usepackage{algpseudocode}
\usepackage{bbding}
\usepackage{pifont}
\usepackage{wasysym}
\usepackage{amssymb}
\usepackage{booktabs}
\usepackage[multiple]{footmisc}
\usepackage[dvipsnames]{xcolor}
\usepackage[export]{adjustbox}
\usepackage{listings}
\usepackage[colorinlistoftodos]{todonotes}
\usepackage{subcaption}
\usepackage{url}
\usepackage{array}
\usepackage{float}

\usetikzlibrary{positioning, fit, calc, arrows.meta, backgrounds}
\newcommand{\xmark}{\ding{55}}

\definecolor{mynavy}{RGB}{0, 0, 128}
\definecolor{mypurple}{RGB}{128, 0, 128}
\definecolor{myturquoise}{RGB}{64, 224, 208}

\definecolor{operationRed}{RGB}{133, 1, 49}
\definecolor{batchedBlue}{RGB}{15, 67, 157}

\DeclareMathOperator*{\argmax}{arg\,max}

\title{\textsc{TreeThink}: A Modular Tree Search Library for \\ Mathematical Reasoning with LLMs} 

\author{Burak S. Akbudak\textsuperscript{$1$}, Zeynel A. Ulusan\textsuperscript{$2,5$}~, Can S. Erer\textsuperscript{$1$}, 
Gözde Gül Şahin\textsuperscript{$2,3,4$}
\\[.3em]
\textsuperscript{1} Computer Engineering Department, Bogazici University, Istanbul, Turkey \\
\textsuperscript{2} Computer Engineering Department, Koç University, Istanbul, Turkey\\
\textsuperscript{3} Friedrich-Alexander-Universität Erlangen-Nürnberg, Intelligent Language Systems \\
\textsuperscript{4} KUIS AI Lab, Istanbul, Turkey \\
\textsuperscript{5} Codeway Studios, Istanbul, Turkey \\
{\url{https://gglab-ku.github.io/}} \\
}

\begin{document}

\maketitle

\begin{abstract}
    Tree search algorithms enable systematic exploration of the proof space in neural theorem proving. Existing LLM tree search libraries primarily target
    natural language reasoning and do not provide native integration with formal
    verifiers, while theorem proving systems often rely on task-specific search
    implementations. We introduce \textsc{TreeThink}, an open-source Python library for modular, fully asynchronous tree search in neural theorem proving. It integrates established tree search methods with vLLM-based inference pipelines and diverse node evaluation techniques, ranging from lightweight heuristics to neural evaluators. We support Lean~4, Rocq, and Isabelle/HOL alongside natural language. It connects directly to each language's Read-Eval-Print Loop (REPL) server for real-time verification and proof state extraction. We evaluate \textsc{TreeThink} on
    miniF2F and MATH500, demonstrating cross-language formal proof search,
    natural language reasoning support, and up to 8.0$\times$ wall-clock speedup
    from asynchronous execution. Source code is released under the MIT license at \url{https://github.com/GGLAB-KU/treethink}, and the library is accessible as a downloadable package at \url{https://pypi.org/project/treethink/}.
\end{abstract}

\section{Introduction}
\label{sec:intro}

\begin{figure}[ht]
    \centering
    \includegraphics[width=\linewidth]{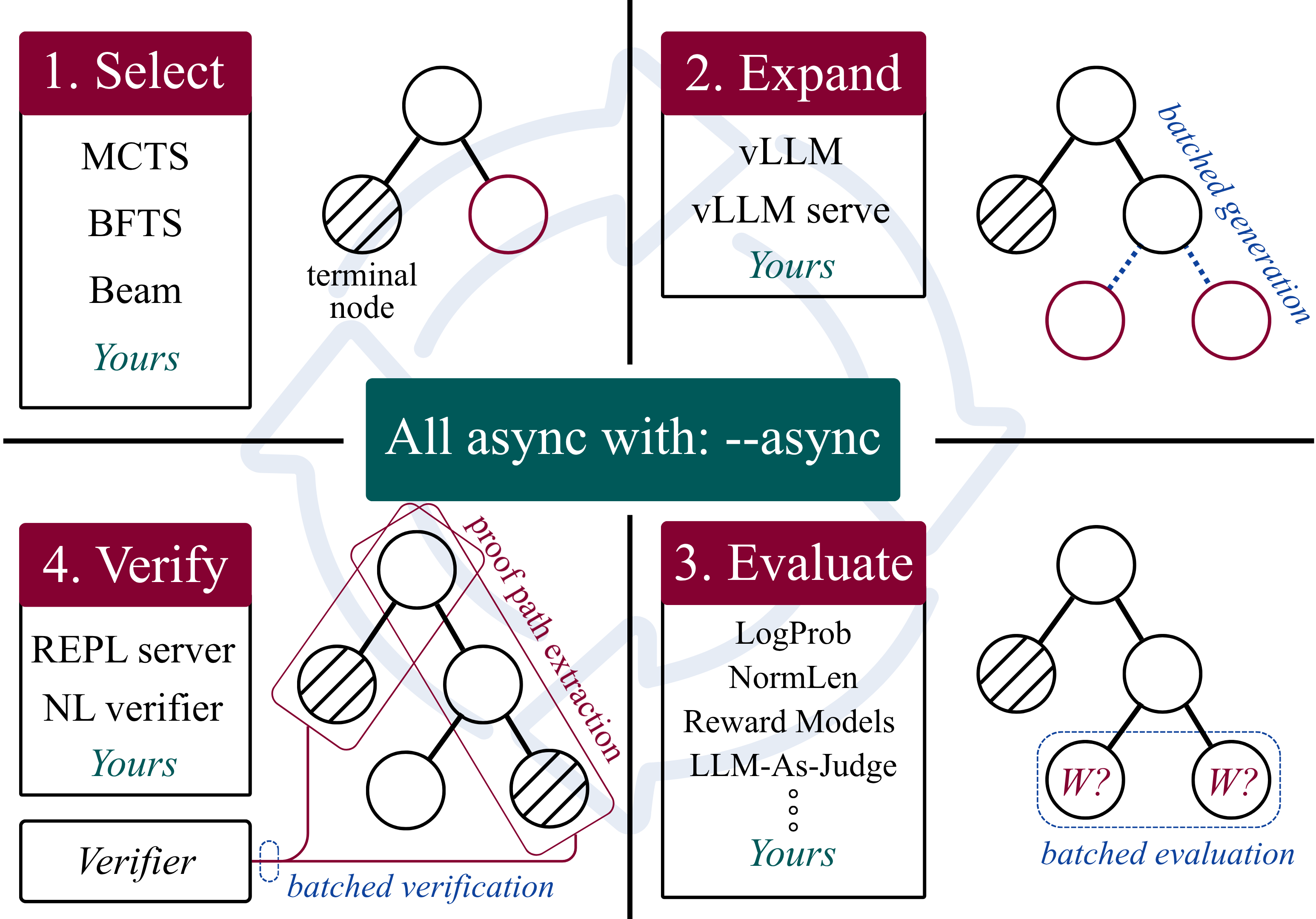}
    \caption{NTP tree search process. \textit{1. Select:} search method selects a node using a search algorithm. \textit{2. Expand}: the policy LLM generates child nodes. \textit{3. Evaluate}: evaluator strategy scores the generated nodes. \textcolor{operationRed}{\textit{W}} stands for the value assigned to a node. \textit{4. Verify}: external systems verify the correctness of the proof. Main operations in individual sections are in \textcolor{operationRed}{red} while batched processes are in \textcolor{batchedBlue}{blue}.}
    \vskip -0.1in
    \label{fig:ntp}
\end{figure}

\begin{table*}[t]
\centering
\scriptsize
\setlength{\tabcolsep}{4pt}
\begin{tabular}{l p{2.0cm} c c p{1.2cm} c c c c}
    \toprule
     & \textbf{Search Methods} & \textbf{\#Policy} & \textbf{\#Evaluators} & \textbf{Language} & \textbf{Formal Verifier} & \textbf{Async.\&Batched} & \textbf{Proof Cache} & \textbf{Graph Vis.} \\
    \midrule
    \texttt{FETCH}         & MCTS, BFS, Beam                & 1 & 1                 & NL & \xmark & \xmark & \xmark & \xmark \\ \addlinespace
    \texttt{LLM Reasoners} & MCTS, BFS, Beam, DFS, CoT, ToT & 6 & \textasciitilde{} & NL & \xmark & \xmark & \xmark & \checkmark\\ \addlinespace
    \texttt{LiTS}          & MCTS, BFS, Beam                & 5 & \textasciitilde{} & NL & \xmark & \xmark & \xmark & \checkmark \\ \addlinespace
    \texttt{Ours}          & MCTS, BFS, Beam                & 2 & 8 & Lean, Rocq, Isabelle, NL & \checkmark & \checkmark & \checkmark & \checkmark \\
    \bottomrule
\end{tabular}
    \caption{Comparison of the proposed framework with existing search libraries. NL stands for natural language. \#Policy: number of supported LLM providers. \#Evaluators: number of node evaluation strategies. Formal Verifier: integration capabilities with formal theorem provers. Proof Cache: presence of proof caching mechanisms. \textasciitilde{} symbol means there exists no implementation that would work in NTP scenario out of the box, and users are required to implement it. }
    \vskip -0.1in
\label{tab:comparison}
\end{table*}

Mathematical reasoning remains a fundamental challenge for artificial intelligence, spanning both informal and formal paradigms \cite{10.1145/3786333}. While large language models (LLMs) have achieved impressive performance in informal mathematical reasoning, they remain susceptible to logical hallucinations and non-monotonic errors \cite{10.1007/978-981-96-7071-0_12}. This has motivated an increasing interest in formal theorem proving, where interactive theorem provers (ITPs) such as Lean~\cite{lean4}, Rocq~\cite{coq}, and Isabelle~\cite{isabelle} verify mathematical statements. Neural theorem proving~(NTP) combines these approaches by using LLMs to propose proof steps while relying on the ITP as a strict verification environment. However, NTP remains a highly difficult problem as the action space of valid mathematical tactics is vast. To guide models into successful trajectories, recent works integrate structured search algorithms into the inference process \cite{li2024a}. Fig.~\ref{fig:ntp} illustrates a typical tree search loop: the search algorithm selects which partial proof to expand, an LLM proposes the next tactic, an evaluator scores nodes, and the proof assistant checks validity. Developing novel proof search systems often requires researchers to reimplement standard execution and verification components, resulting in substantial engineering overhead. For instance, to show the effectiveness of their proposed proof level reward evaluator, \citealp{wang-etal-2023-dt} implement a tree search mechanism from the ground up, and \citealp{hunyuanprover} orchestrate a search system to evaluate the proposed process reward models. General LLM tree search libraries
such as LLM Reasoners~\cite{hao2024llm}, FETCH~\cite{fetch}, and
LiTS~\cite{li-tao-2026-lits} provide reusable search abstractions, however users
must still implement formal proof verification, proof-state extraction, and
domain-specific evaluators for NTP.

To reduce duplication of effort and accelerate experimentation, we present
\textsc{TreeThink}, a modular Python library for tree search in neural theorem
proving. \textsc{TreeThink} is designed
around formal verifier interaction with natural language
support. Our contributions are threefold: \textsc{TreeThink} decouples search
algorithms, LLM policies, evaluators, and environment interaction into reusable
components; provides unified REPL clients for Lean~4, Rocq, and Isabelle/HOL;
and supports fully asynchronous and batched execution with vLLM for high
throughput inference~\cite{kwon2023vllm}. We evaluate the system on miniF2F
for formal mathematics~\cite{minif2f} and MATH500 for natural language
mathematical reasoning~\cite{lightman2023lets}, showing support for both
settings and up to 8.0$\times$ speedup over synchronous execution. Alongside the open source code\footnote{https://github.com/GGLAB-KU/treethink}, we release the framework as a downloadable package in PyPI\footnote{https://pypi.org/project/treethink/}, and provide a demo video\footnote{https://youtu.be/vKFXxnjlk8M}.

\section{Previous Systems} 
\label{sec:prev}

Recent efforts in LLM-based reasoning increasingly treat tree search as a reusable infrastructure by separating search logic from model inference~\cite{fetch,hao2024llm,li-tao-2026-lits}. These systems provide useful
abstractions for natural-language reasoning, but they do not natively integrate
formal proof verification, proof-state extraction, or multi-ITP execution. In
formal reasoning, existing tools typically address individual parts of the
pipeline:  extract proof state information~\cite{yang2023leandojo}, execute proofs~\cite{pantograph}, or support premise retrieval~\cite{leansearchv2}. Furthermore, recent formal math theorem-proving systems~\cite{xin-etal-2025-bfs, hunyuanprover, deepseek-prover-v15, wu2025internlm25stepprover} are 
designed as task-specific prover pipelines rather than reusable, multi-language
tree-search libraries. Thus, existing systems either provide a general-purpose LLM tree search without formal verifier integration, or the provided infrastructure lacks reusable, multi-language, asynchronous tree-search abstractions. \textsc{TreeThink} fills this gap by combining reusable tree-search components, batched LLM inference, heterogeneous evaluators, and unified REPL interaction across Lean, Rocq, and Isabelle/HOL. We compare our framework with other general tree search systems in Table~\ref{tab:comparison}. 

\section{\textsc{TreeThink}}
\label{sec:treethink}

\begin{figure*}[t]
    \makebox[\textwidth][c]{%
        \resizebox{0.9\textwidth}{!}{%
            \begin{tikzpicture}[
                >=Stealth,
                base/.style={
                    rectangle, ultra thick, rounded corners=2mm, 
                    minimum width=3.8cm, minimum height=1.5cm, align=center
                },
                inputnode/.style={
                    base, draw=cyan!70!black, fill=cyan!5
                },
                corenode/.style={
                    base, draw=teal!70!black, fill=teal!5
                },
                helpernode/.style={
                    base, draw=orange!80!black, fill=orange!10
                },
                outsidenode/.style={
                    base, draw=purple!70!black, fill=purple!5
                },
                outerbox/.style={
                    rectangle, draw=gray!70, thick, fill=gray!5, 
                    rounded corners=4mm, inner sep=4mm
                },
                innerbox/.style={
                    rectangle, draw=gray!70, thick, fill=white, 
                    rounded corners=4mm, inner sep=4mm
                }
            ]

            \node[corenode] (search) {Search Method \\ {\color{black!80}\footnotesize Main orchestrator}};
            \node[corenode] (eval) [left=1.5cm of search, yshift=1.2cm] {Evaluator \\ {\color{black!80}\footnotesize Node scoring}};
            \node[corenode] (policy) [left=1.5cm of search, yshift=-1.2cm] {Policy \\ {\color{black!80}\footnotesize Node generator}};
            \node[corenode] (replclient) [above=1.0cm of search, xshift=-0.0cm] {REPL Client \\ {\color{black!80}\footnotesize Environment Interaction}};
            
            \node[helpernode] (replruntime) [right=1.5cm of search] {REPL Runtime \\ {\color{black!80}\footnotesize REPL orchestrator}};
            \node[helpernode] (proof) [below=0.75cm of replruntime] {Proof Cache \\ {\color{black!80}\footnotesize Checked proof set}};
            
            \begin{scope}[on background layer]
                \coordinate (tt_top) at ([yshift=1.3cm]eval.north);
                \coordinate (tt_bottom) at ([yshift=-0.3cm]policy.south);
                \coordinate (tt_left) at ([xshift=-0.3cm]eval.west);
                \coordinate (tt_right) at ([xshift=0.3cm]search.east);
                
                \coordinate (bs_top) at ([yshift=1.2cm]replclient.north);
                \coordinate (bs_bottom) at ([yshift=-0.0cm]proof.south);
                \coordinate (bs_left) at ([xshift=-0.6cm]tt_left);
                \coordinate (bs_right) at ([xshift=0.2cm]replruntime.east);
                
                \node[outerbox, fit=(bs_top) (bs_bottom) (bs_left) (bs_right)] (sampler) {};
                \node[anchor=north west, xshift=4mm, yshift=-4mm, text=gray!80!black] 
                    at (sampler.north west) {\large Batched Sampler};

                \node[innerbox, fit=(tt_top) (tt_bottom) (tt_left) (tt_right)] (treethink) {};
                \node[anchor=south east, xshift=-4mm, yshift=4mm, text=gray!80!black] 
                    at (treethink.south east) {TreeThink};
            \end{scope}

            \node[outsidenode] (replserver) [right=1.2cm of replruntime] {REPL server \\ {\color{black!80}\footnotesize Formal proof verification}};
            \node[outsidenode] (vllm) [below=2.0cm of policy] {vLLM \\ {\color{black!80}\footnotesize Inference engine}};

            \node[inputnode, minimum width=4cm, minimum height=1.5cm, rotate=90, text width=3.6cm, align=center] (cli) at ([xshift=-1.2cm]sampler.west |- search) 
                {CLI \& Parsing \\ {\color{black!80}\footnotesize Input pipeline}};
            \node[inputnode] (yaml) at ([xshift=-3.2cm, yshift=1cm]cli.center) {YAML Configuration \\ {\color{black!80}\footnotesize System settings}};
            \node[inputnode] (data) at ([xshift=-3.2cm, yshift=-1cm]cli.center) {Datasets Registry \\ {\color{black!80}\footnotesize Dataset related information}};

            \draw[->, thick, draw=gray!80!black] (yaml.east) -- (cli.north |- yaml.east);
            \draw[->, thick, draw=gray!80!black] (data.east) -- (cli.north |- data.east);
            \draw[->, thick, draw=gray!80!black] (cli.south) -- (sampler.west |- cli.south);

            \draw[->, thick, draw=gray!80!black] (eval.east) -- (search.west);
            \draw[->, thick, draw=gray!80!black] (policy.east) -- (search.west);

            \draw[->, thick, draw=gray!80!black] (replclient.south) -- (search.north);
            \draw[->, dashed, thick, draw=gray!80!black, rounded corners=2mm] (replclient.west) -| (eval.north);
            \draw[->, thick, draw=gray!80!black] (search.east) -- (replruntime.west);
            \draw[<->, thick, draw=gray!80!black] (replclient.east) -- ++(3.4, 0.0) -- (replruntime.north);

            \draw[<->, thick, draw=gray!80!black] (replruntime.east) -- (replserver.west);
            \draw[<->, dashed, thick, draw=gray!80!black] (replruntime.south) -- (proof.north);
            
            \draw[->, thick, draw=gray!80!black] (policy.south) -- (vllm.north);
            \draw[->, dashed, thick, draw=gray!80!black, rounded corners=2mm] 
                (eval.west) -- ++(-0.3,0) |- (vllm.west);

            \end{tikzpicture}%
        }%
    }
    \centering
    {\footnotesize
    {\color{cyan!70!black}\rule{1.2ex}{1.2ex}} Input pipeline \;
    {\color{teal!70!black}\rule{1.2ex}{1.2ex}} Core components \;
    {\color{orange!80!black}\rule{1.2ex}{1.2ex}} Helper units \;
    {\color{purple!70!black}\rule{1.2ex}{1.2ex}} External system
    }
    \caption{Overview of the system. Our framework accepts search configuration as a YAML file and a dataset registry. Then, we parse the given parameters and initiate search components. We embed the environment interaction logic into helper classes for modularity. Dashed lines denote optional relationships as some features may require the indicated connection under specific configurations (e.g., when the evaluator is selected as neural model).}
    \label{fig:overview}
\end{figure*}
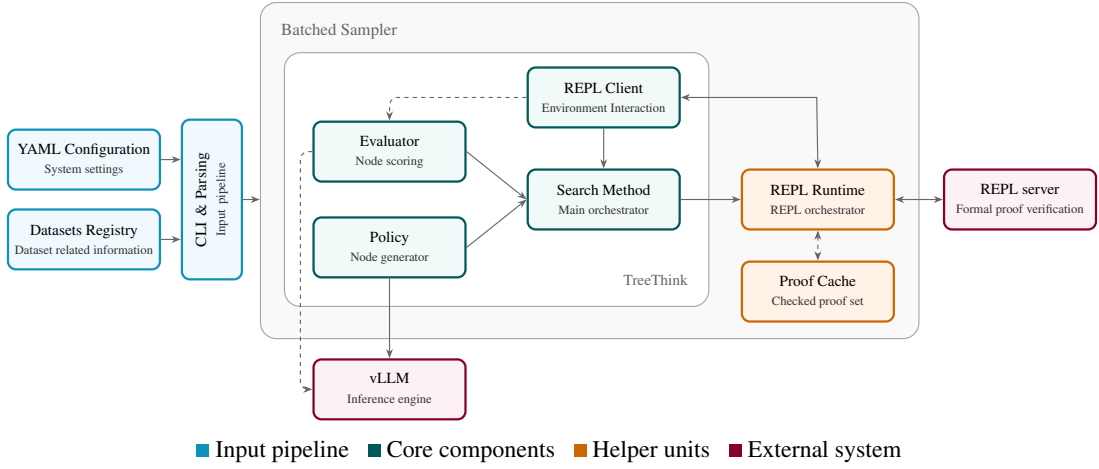

\textsc{TreeThink} is designed around three principles: component interchangeability, verifier-agnostic environment interaction, and asynchronous execution. Component interchangeability is achieved by modularizing search methods~\S\ref{sec:search}, candidate-generating policies~\S\ref{sec:policies}, and step evaluators~\S\ref{sec:evaluators}. Verifier-agnostic interaction is managed by environment interactors~\S\ref{sec:environment}, which standardize communication with formal proof checkers and informal verifiers. All components support asynchronous, batched execution. In addition, \textsc{TreeThink} incorporates several features to accelerate experimentation~\S\ref{sec:add-feat}. An overview is shown in Figure~\ref{fig:overview}.

\subsection{Search Algorithms}
\label{sec:search}

Search algorithms construct the proof tree by iteratively selecting and expanding nodes via a policy model, guided by leaf value estimates from evaluators. Each algorithm features an asynchronous counterpart to enable concurrent selection, expansion, and evaluation. We adapt four algorithms with varying exploration behavior and computational cost that have demonstrated effectiveness in recent NTP studies \cite{polu2020generative, xin-etal-2025-bfs, wu2025internlm25stepprover, DBLP:journals/corr/abs-2402-03300}.



\paragraph{Best-First Search} \cite{Pearl1984HeuristicsI} aims to prioritize the most promising frontier nodes according to evaluator scores. It does so by maintaining a priority queue of leaf nodes that are eligible for expansion and expanding the most promising node at each iteration:
\[
s^* = \argmax_{s \in \mathcal{F}} V(s)
\]
where $\mathcal{F}$ is the frontier of leaf nodes and V(s) is the previously assigned evaluator score. The algorithm expands the selected node by generating candidate tactics and evaluates the generated states using the specified evaluation strategy.

\paragraph{Beam Search} \cite{bisiani1987beam} focuses on pruning low-quality branches early. The algorithm expands the proof tree level-by-level, keeping the top-$k$ most promising states at each depth:
\[
\mathcal{B}_{d+1} = \text{top-}k\left(\bigcup_{s \in \mathcal{B}_d} 
\text{children}(s)\right)
\]
where states are ranked by evaluator scores. At each level, we expand all beam states in parallel, generating multiple candidates per state. The top-$k$ children across all expansions form the next beam, i.e., parent nodes do not survive.

\paragraph{Traditional Monte Carlo Tree Search (MCTS)} \cite{mcts} systematically explores search spaces by balancing exploration and exploitation. Each iteration involves selection, expansion, simulation, and backpropagation. During selection, the tree is traversed using the Upper Confidence Bounds for Trees (UCT) formula:
$$s^* = \argmax_{s_i \in \mathrm{child}(s)} \left[Q(s_i) + c\sqrt{\frac{\ln N(s)}{N(s_i)}}\right]$$
where $Q(s_i)$ is the mean value estimate of child $s_i$, $N$ denotes visit counts, and $c$ controls exploration. The policy expands the state into new child nodes, from which $n$ terminal rollouts are generated. An evaluator scores these rollouts, assigning the mean value $V$ to the child node. Finally, values are backpropagated up to the root via $Q(s) \leftarrow \frac{N(s) \cdot Q(s) + V}{N(s) + 1}$.

\paragraph{Rollout-Free Value-Guided MCTS}
Traditional simulation is computationally expensive in NTP as each node requires terminal rollouts. Therefore, we implement a rollout-free variant inspired by AlphaZero \cite{silver2017alphazero}, replacing simulation with direct value evaluation. The algorithm selects an expandable node using the aforementioned UCT rule and generates candidate proof steps. \textsc{TreeThink} then directly evaluates these candidates to produce a value estimate, which is backpropagated to update ancestor nodes. Unlike AlphaZero, \textsc{TreeThink} utilizes generic UCT rather than PUCT with learned priors. This accommodates arbitrary pluggable evaluators better suiting the computational constraints of NTP.

\subsection{Policies}
\label{sec:policies}


Policies enlarge the search tree by generating child nodes from LLM outputs. \textsc{TreeThink} supports local and remote inference via vLLM's \texttt{LLM} and \texttt{vllm serve} interfaces respectively. We use \texttt{AsyncLLM} for handling concurrent generation across search branches. When the policy is called, the model is prompted with the current proof trajectory starting from the root node to selected child node. Full model outputs are stored in each node, allowing heuristic evaluators such as \texttt{LogprobEvaluator} to leverage it. Output text is parsed in two modes: newline-delimited (default) and XML-tag-delimited. In newline-delimited mode, we give the newline token as the stop token to vLLM's sampling parameters. For XML mode, users specify a tag such as \texttt{\textless{}PROOF\_STEP\textgreater{}}; we set the corresponding closing tag as the stop token and extract the inner content to construct nodes. Using XML tags allows LLMs to separate reasoning from proof steps independent of the selected search components. With this approach, models that produce more structured outputs can be natively integrated into our framework. Optionally, \textsc{TreeThink} deduplicates child nodes by comparing their text before adding them to the search tree. This eliminates redundant candidates, conserving the expansion budget and ensuring each expansion yields a distinct exploration path.


\subsection{Evaluators}
\label{sec:evaluators}


Evaluators score generated nodes to determine the search algorithm's expansion priorities (\S\ref{sec:search}). We implement adaptable approaches established in the literature. Asynchronous complements for heuristic evaluators wrap synchronous implementations, while neural reward systems leverage vLLM's asynchronous capabilities. Table~\ref{tab:evaluator_summary} summarizes the supported evaluators detailed below:


\noindent \texttt{LogprobEvaluator} scores nodes by cumulative log-probability of the generated text within that individual node, favoring more likely reasoning steps. This technique is a fundamental approach used by GPT-f \cite{polu2020generative}.

\noindent \texttt{NormLenEvaluator} pioneered by \citealp{xin-etal-2025-bfs}, applies a length-normalized scoring function, 
\[
V(s) = \frac{\Sigma_{t=0}^{L-1} \log p(a_t | s_t)}{L^{\alpha}}
\]
where $L$ is node depth, $\alpha$ controls the strength of the length penalty, $s_t$ and $a_t$ are proof step and applied tactic at time $t$ respectively, and $p(a_t | s_t)$ the predicted probability of model generating the next tactic $a_t$ at state $s_t$. Normalizing by $L^\alpha$ enables more fine-grained control over the shape of the tree, e.g., increasing $\alpha$ would lead the tree into deeper proof paths. 


\noindent \texttt{ProofLevelRewardEvaluator} scores a node by evaluating the complete generated proof with a discriminative reward model via vLLM pooling mode. The pooling operation can be either ``classify'' for sequence reward models outputting a scalar value, or ``token\_classify'' for process reward models (PRM) producing a per-token score, which can be reduced to a single scalar by taking the mean. The final score reflects how promising the partial proof is, judged in the context of the entire proof rather than a single step. \citealp{wang-etal-2023-dt} use this approach in a formal proof search setting with their DT-Solver system.

\begin{table}[t]
\centering
\scriptsize
\begin{tabular}{l c c c c c}
    \toprule
    \textbf{Evaluator} & \textbf{Signal Type} & \textbf{REPL?} & \textbf{RM?}  \\
    \midrule
    \texttt{LogprobEvaluator}          & Continuous & \xmark     & \xmark      \\
    \texttt{NormLenEvaluator}          & Continuous & \xmark     & \xmark      \\
    \texttt{ProofLevelRewardEvaluator} & Discrete   & \xmark     & \checkmark  \\
    \texttt{StateLevelRewardEvaluator} & Discrete    & \xmark     & \checkmark \\
    \texttt{JudgeEvaluator}            & Continuous & \checkmark & \checkmark \\
    \texttt{TournamentEvaluator}       & Continuous & \checkmark & \checkmark \\
    \texttt{RMaxTSEvaluator}           & Continuous & \xmark     & \xmark \\
    \texttt{REPLEvaluator}             & Binary     & \checkmark & \xmark \\
    \bottomrule
\end{tabular}
\caption{Supported evaluators with their signal type, REPL requirement, and neural reward model (RM) usage. Due to the modular design, all evaluators are suitable for both informal and formal math problem solving.
}
\vskip -0.1in
\label{tab:evaluator_summary}
\end{table}

\noindent \texttt{StateLevelRewardEvaluator} is similar to the proof-level variant but uses only the current node's text rather than the full proof, acting as a PRM. It judges the sensibility of an individual step independent of history. \citealp{zhang2024restmcts} uses a similar evaluator system in their self-training pipeline.

\noindent \texttt{JudgeEvaluator} utilizes a REPL to extract state information and a generative reward model (GRM) that assigns quality scores to proof steps. The model is prompted with current goals, applied tactic, and solved goals, and is asked to give a score on a fixed scale (e.g. 0–20). Given score is then parsed, and normalized to $[0, 1]$. We can see this evaluator in action in STILL \cite{still1} framework, where authors use a GRM.



\noindent \texttt{TournamentEvaluator} ranks candidate proof steps using a single-elimination tournament. After extracting proof states via batched REPL calls, a separate LLM judge performs batched pairwise comparisons. Candidates are scored based on their elimination round, with the overall winner receiving the total number of rounds. Normalizing these scores to $[0,1]$ yields a continuous quality signal reflecting tournament advancement. \citealp{mahdavi2026scaling} employ a similar pairwise strategy for proof-level selection.

\noindent \texttt{RMaxTSEvaluator} designed by \citealp{deepseek-prover-v15}, assigns a novelty bonus to previously unseen proof states, which are tracked via SHA-256 hashing. When extrinsic rewards from proof verification are sparse, it gives the search a continuous incentive to explore new parts of the proof space.

\noindent \texttt{REPLEvaluator} returns binary type-check feedback by executing proof steps in a formal REPL. It is mainly intended for traditional MCTS rollouts where binary signals suffice.

\subsection{Environment Interaction}
\label{sec:environment}



We implement a unified client interface for communicating with formal proof checkers' REPL servers: i) Kimina Lean Server~\cite{santos2025kiminaleanserver} for Lean~4  by adapting its existing sync/async clients, ii) isabelle-server~\cite{isabelle} via isabelle-client~\cite{isabelleclient} for Isabelle, and iii) \texttt{rocq-ml-server} from rocq-ml-toolbox~\cite{rocqmltoolbox} for Rocq. Both rocq-ml-server and isabelle-server accept proof paths rather than raw strings. Therefore, we provide proofs inside temporary files that are cleaned up after use, incurring a negligible I/O overhead. To batch the inputs and allow concurrency, we connect to servers via multiple clients.

\textsc{TreeThink} supports three modes of formal-language REPL interaction: (1) evaluator-driven interactions, where evaluators query the REPL during the search process (e.g., JudgeEvaluator); (2) post-search verification, which validates completed proof candidates after search termination; and (3) online verification, which evaluates candidates at terminal nodes (e.g. those with triple ticks as their text) during search to enable early stopping. All verification methods support batched execution to balance accuracy against runtime overhead.

\subsection{Additional Features}
\label{sec:add-feat}

Beyond core components, \textsc{TreeThink} includes features that improve search speed, enable visual inspection, and simplify usage.

\paragraph{Proof Caching} An LRU cache keyed by SHA-256 hash of proof text prevents duplicate REPL submissions. Proof caching is especially important in tree search with NTP as branches often produce duplicate terminal proof candidates.

\paragraph{Graph Interaction} Graphviz-based \cite{Ellson2004} visualizations with auto-computed statistics facilitate visual debugging of search components. Additionally, search states can be saved and reloaded to resume interrupted experiments.

\paragraph{Configuration and Execution} Dataset metadata is stored in TOML \cite{toml-lang} files to ensure reproducibility, while experiments run via a unified command-line interface that accepts all search configurations, making the framework well-suited for HPC systems (see Appendices~\ref{sec:dataset_registry} and \ref{sec:cmd}).

\section{Evaluation of the System}
\label{sec:evaluation}

Our evaluation approach consists of three stages: cross-language formal proof search~\S\ref{sec:eval-formal} where we illustrate our framework's capability of interacting with different ITPs, comparison between asynchronous and synchronous runs~\S\ref{sec:eval-async} to observe the gained speedups, and informal proof search with fixed-budget~\S\ref{sec:eval-informal} to showcase the effectiveness of tree search. We demonstrate the system's modularity in Appendix~\ref{sec:eval-modularity} by detailing the lines of code required to switch languages. Additionally, Appendix~\ref{sec:eval-coverage} outlines the smoke tests used to validate approaches outside our main evaluation. 

Our aim with these experiments is to demonstrate our framework's ability to act as a reusable tree-search system rather than a SOTA theorem proving system. Unless otherwise stated, the default experimental configuration uses a search budget of 512 expansions with 3 children per expansion, a vLLM policy backend, a NormLenEvaluator formal evaluator, the miniF2F test and MATH500 as formal and informal datasets respectively, and a single NVIDIA A40 GPU. We provide a comprehensive list of experiment parameters in Appendix~\ref{sec:reproducibility} for reproducibility purposes.



\subsection{Cross-Language Formal Proof Search}
\label{sec:eval-formal}


We evaluate our system on formal reasoning using the miniF2F dataset, which serves as an appropriate benchmark for cross-language evaluation, as it is available in all three formal languages that TreeThink supports. We use DeepSeekProverV2-7B \cite{ren2025deepseekproverv2} for Lean, Qwen2.5-Coder-14B-Instruct \cite{hui2024qwen2} for Isabelle, and Qwen3.5-9B \cite{qwen3.5} for Rocq for our runs. Since model availability differs across formal languages, we select a strong publicly available model for each language, using prior work when available~\cite{hybridprover}.




\begin{table}[t]
    \centering
    \footnotesize
    \setlength{\tabcolsep}{4pt}
    \renewcommand{\arraystretch}{0.95}
    \begin{tabular}{@{}lccc@{}}
        \toprule
        Method & Lean (\%) & Isab. (\%) & Rocq (\%) \\
        \midrule
        pass@1  & 49.4 & 42.2 & 0.8 \\
        BFTS    & 54.9 & 65.1 & 1.2 \\
        RF-MCTS & 57.4 & 61.8 & 2.0 \\
        \bottomrule
    \end{tabular}
    \caption{
    Formal proof-search pass rates on miniF2F dataset. RF-MCTS denotes our rollout-free value-guided MCTS variant. Tree-search methods use the same expansion budget, child budget, policy backend, and evaluator.
    }
    \label{tab:minif2f_formal}
    \vskip -0.1in
\end{table}

Table~\ref{tab:minif2f_formal} shows tree search consistently improves pass@1 rates across all three languages. RF-MCTS outperforms BFTS in Lean and Rocq, while the reverse holds for Isabelle, demonstrating the value of modular search methods. The overall performance in Rocq remains comparatively low, which we attribute to the limited availability of Rocq-specialized public models and prior benchmarks.

\subsection{Asynchronous Execution}
\label{sec:eval-async}


We evaluate the asynchronous features by varying the concurrency level, with pass rate and average wall-clock time per question results presented in Table~\ref{tab:async_concurrency}. Across all runs, we maintain Isabelle as the formal language, and Qwen2.5-Coder-14B-Instruct as the policy model. We aim to demonstrate how concurrency affects the average solution time, rather than an improvement in accuracy. For demonstration purposes, we use a subset of miniF2F with 32 random samples.

\begin{table}[h]
    \centering
    \footnotesize
    \setlength{\tabcolsep}{3.5pt}
    \renewcommand{\arraystretch}{0.95}
    \begin{tabular}{@{}lrrr@{}}
        \toprule
        Mode / Concur. & Pass (n=32) & Avg. Time / Q. & Speedup\\
        \midrule
        Sync / 1  & 18.50 & 3.72m & x1.00 \\
        Async / 2 & 17.75 & 2.42m & x1.53 \\
        Async / 4 & 17.00 & 1.05m & x3.54 \\
        Async / 8 & 17.75 & 0.74m & x5.02 \\
        Async / 16 & 16.25 & 0.46m & x8.08 \\
        \bottomrule
    \end{tabular}
    \caption{
    Effect of asynchronous execution and concurrency level. All settings use
    the same model, evaluator, expansion and child budget. Avg. Time/Q.
    denotes average wall-clock time per question. We run each Mode / Concurrency experiment four times and report the average pass rates and Avg. Time/Q.
    }
    \label{tab:async_concurrency}
    \vskip -0.1in
\end{table}




As shown in Table~\ref{tab:async_concurrency}, asynchronous runs significantly reduces the wall-clock time at a small cost of accuracy. Specifically, we achieve 8.0$\times$ speedup when concurrency level is set to 16.



\subsection{Informal Mathematical Reasoning}
\label{sec:eval-informal}


To demonstrate our framework's support for natural language mathematical reasoning, we use MATH500, a subset of the MATH~\cite{hendrycksmath2021} dataset. It contains challenging competition-level problems stated in natural language. We select LLama3-8B \cite{grattafiori2024llama3herdmodels} as our policy model. For the evaluator, we use \texttt{JudgeEvaluator} with the judge model as RISE-Judge-Qwen2.5-7B \cite{yu2025improvellmasajudgeabilitygeneral}. 

To compare the models' answers with the ground truth (GT), we prompt the model to provide the final answer inside \verb|\boxed{}|. After extracting the contents, we parse the text into expressions compatible with the symbolic mathematics library SymPy~\cite{sympy} to verify equality between GT and the proposed solution.

\begin{table}[h]
    \centering
    \footnotesize
    \setlength{\tabcolsep}{4pt}
    \renewcommand{\arraystretch}{0.95}
    \begin{tabular}{@{}lr@{}}
        \toprule
        Method & Pass (\%) \\
        \midrule
        pass@1 & 23.0 \\
        maj@12 & 34.8 \\
        RF-MCTS & 40.0 \\
        \bottomrule
    \end{tabular}
    \caption{
    Natural-language mathematical reasoning results on MATH500. RF-MCTS indicates rollout-free value-guided MCTS with a judge-based evaluator.
    }
    \label{tab:math500_results}
    \vskip -0.1in
\end{table}


We report pass rates for pass@1, majority voting@12 \cite{wang2023selfconsistency} and our tree search run on MATH500 at Table~\ref{tab:math500_results}. We choose maj@12, since RF-MCTS produces approximately
12 terminated paths per problem on average. Among tested methods, RF-MCTS achieves the highest score, suggesting that TreeThink's search abstraction can also benefit natural language mathematical reasoning. Moreover, this experiment demonstrates our framework's capability to model natural language reasoning without structural changes.

\section{Conclusion}
\label{sec:conc}
We present \textsc{TreeThink}, a modular, open-source library for tree search in neural theorem proving. By decoupling infrastructure into reusable search algorithms, evaluators, and a vLLM-backed policy layer, it eliminates significant engineering overhead. Its fully asynchronous architecture yields substantial wall-clock speedups without accuracy loss. Furthermore, a unified REPL client enables identical search configurations across three ITPs. We also test the support for natural language reasoning, and achieve higher results compared to single-pass inference. We believe that TreeThink's extensible design simplifies the integration of new components, providing a practical foundation for future research in formal and informal reasoning.


\section*{Limitations}

\paragraph{Evaluation strategy} Our evaluation showcases the framework's capabilities; large-scale benchmarking across diverse datasets and model families is left for future work.

\paragraph{Inference provider support} \textsc{TreeThink} currently supports only vLLM-based inference (local or \texttt{vllm serve}). Other backends are not yet integrated, though the policy abstraction can accommodate them via subclassing.

\paragraph{Tool-assisted reasoning} \textsc{TreeThink} does not yet support interleaving proof steps with external tool calls (e.g., symbolic computation). Incorporating tool-use into the search loop is left as future work.

\paragraph{Domain scope} The library is currently tailored to mathematical theorem proving. Extending the framework to other reasoning domains such as code generation, agentic task planning, or scientific discovery would require additional environment interactors and domain-specific evaluation strategies.

\paragraph{Search visualization} While the framework provides Graphviz-based static visualizations of completed search trees, it does not support animated or interactive exploration of the search process (e.g., step-by-step playback or live tree expansion in a browser interface).

\paragraph{Component benchmarking}
\textsc{TreeThink} supports multiple search methods and evaluators, but our main
experiments evaluate only representative configurations. A full factorial
benchmark over all search methods, evaluators, models, and proof assistants is
computationally expensive and left for future work.

\section*{Broader Impact Statement}
By allowing researchers to easily extend its capabilities with new search methods, evaluators, policies, or formal languages, TreeThink lowers the engineering barriers to automated formal reasoning research. It provides mathematicians and researchers in neural theorem proving with accessible, LLM-driven tools to explore complex proof spaces, aiding the advancement of mathematical understanding. Furthermore, the library's asynchronous and batched execution reduce the computational overhead and environmental footprint typically associated with large-scale reasoning tasks.

\section*{Acknowledgements}
This work has been partially supported by the Scientific and Technological Research Council of Türkiye (TÜBİTAK) as part of the project “Automatic Learning of Procedural Language from Natural Language Instructions for Intelligent Assistance” with the number 121C132. We also gratefully acknowledge KUIS AI Lab for providing computational support.

\bibliographystyle{acl_natbib}
\bibliography{custom}

\appendix
\clearpage

\section*{Appendix} 
\label{sec:appendix}

\section{Full List of Controllable Parameters}
\label{sec:params_list}

We provide the full list of controllable parameters in our framework below. Note that, REPL client, model and sampling parameters are not given multiple times as they share the same dataclass.

\begin{minted}[
    frame=lines,
    framesep=1.5mm,
    baselinestretch=1.0,
    fontsize=\footnotesize,
    linenos,
    numbersep=3pt
]{yaml}
treethink:
  method_name: "BFTS"
  max_children: 3
  expansion_count: 512
  timeout: 480
  graph_path: /path/to/graph/output
  termination_str: "```"
  store_method_class: false
  store_graph_stats: true
  remove_duplicate_children: true
  parse_tag: "\n"
    
  # REPL / termination
  language: "lean"
  client_args:
    host: "localhost"
    port: "5000"
    batch_size: 8
    num_proc: 4
    timeout: 200
    enable_cache: true
    cache_maxsize: 4096
  termination_on_encounter:
      batch_size: 2
  termination_on_paths:
      max_repl: 16
  max_concurrent_expansions: 8
    
  # Special to Search Methods 
  beam_width: 4
  exploration_weight: 1.414
  final_decision_mode: "native"
  tie_breaker: "random"
  
  # Rollout (TraditionalMCTS)
  rollout_evaluator_args: null
  rollout_max_tokens: 4096
  rollout_n: 1
  rollout_temperature: 1.0
  rollout_top_p: 0.9

policy:
  func_name: "vllm_policy"
  model: 
    model: "deepseek-ai/DeepSeek-Prover-V2-7B"
    max_model_len:
    tensor_parallel_size: 1
    gpu_memory_utilization: 0.95 
  sampling:
    max_tokens: 2048
    temperature: 1.0
    top_k: -1
    top_p: 0.95
    seed: 
    stop: ["\n"]  
    n: 3
    logprobs: 1
  server: # vLLM server mode
    base_url: "http://localhost:8000/v1"
    api_key: ""
    timeout: 600
  visible_devices: "0"

evaluator:
  func_name: "cumulative_logprob_evaluator"
  length_norm: 0.5 # 
  client_args: # client args as before

  # for JudgeEvaluator
  llm_as_judge_model: # model args as before
  llm_as_judge_sampling: # sampling args as before
  llm_as_judge_system_prompt: "You are 
  an LLM judge..."
  llm_as_judge_visible_devices: "1"
  
  # for Discriminative reward models
  reward_model: # model args as before
  reward_pooling_task: "classify"
  reward_score_reduction: "mean"
  reward_system_prompt: "Give reward based on..."
  reward_visible_devices: "0"
\end{minted}

\section{Example Dataset Registry Entry}
\label{sec:dataset_registry}
We implement a simple dataset registry for enhanced reproducibility in our experiments. An entry includes a name, dataset path, system and user prompts, what data key to use and its renamed counterparts for compatibility with the framework. An example dataset registry is given below:

\begin{minted}[
    frame=lines,
    framesep=1.5mm,
    baselinestretch=1.0,
    fontsize=\footnotesize,
    linenos,
    numbersep=4pt
]{toml}
[configs.minif2f_lean]
name="minif2f_lean"
path_or_name="/path/to/minif2f/lean/dataset"
dataset_split="test"
prompt_format="""Complete the following lean 
code: \n```\n{}"""
system_prompt = "You are an expert in Lean 4 
formal proof language."
data_keys=["formal_with_headers", "problem_id"]
renamed_data_keys=["problem", "problem_id"]
format_type="huggingface_disk"
\end{minted}

\section{\texttt{treethink run} Command Example}
\label{sec:cmd}
Below is an example command for running a tree search using our framework:
\begin{minted}[
    frame=lines,
    framesep=1.5mm,
    baselinestretch=1.0,
    fontsize=\footnotesize,
    linenos
]{bash}
treethink run \
  --gen-config-path "search_params.yaml" \
  --data-config-path "dataset_config.yaml" \
  --data-config-name "minif2f_lean" \
  --batch-size 8 \
  --num-iterations 1 \
  --output-dir "/output_folder/" \
  --run-name "minif2f_lean" \
  -v debug \
  --continue-from-prev \
  --async # auto-converts all components to 
          # async implementations
\end{minted}

\section{Further Discussion on System Evaluation}
\subsection{Modularity}
\label{sec:eval-modularity}


\begin{table}[h]
\centering
\footnotesize
\begin{tabular}{@{}ll@{}}
\toprule
Classification    & Components \\
\midrule
Language-Agnostic & Search method, Evaluator,  \\
                  & Policy, Search\&Child budget \\
\addlinespace
Language-Specific & REPL client, Policy model,  \\
                  & Prompt template \\
\bottomrule
\end{tabular}
\caption{Classification of TreeThink's components in terms of language dependency. Notably, REPL client is not choosable, as it is determined upon selecting the language.}
\label{tab:modularity}
\vskip -0.1in
\end{table}

As detailed in Section~\ref{sec:treethink}, the modular architecture of \textsc{TreeThink} allows individual components to be swapped independently, facilitating highly reusable search processes. For example, migrating a tree search from Lean to Isabelle requires modifying only four lines in configuration files: language name, selected policy model, prompt and the dataset used. To clarify this separation of concerns, Table~\ref{tab:modularity} categorizes all components as either language-dependent or language-agnostic.

\subsection{Component Coverage}
\label{sec:eval-coverage}


While we do not exhaustively evaluate every combination of search method, evaluator, and policy, our experiments focus on a representative subset that demonstrates the core functionality of our framework. To validate the remaining components, we employ smoke testing, i.e. successful end-to-end execution on a small held-out set shown in Table~\ref{tab:component_coverage}.

\begin{table}[H]
\centering
\footnotesize
\renewcommand{\arraystretch}{1.2}
\begin{tabular}{@{}l p{0.75\linewidth}@{}}
\toprule
Scope & Covered Components \\
\midrule
Main eval. & Best-first search, RF-MCTS, NormLenEvaluator, JudgeEvaluator \\
Smoke test & Best-first search, Beam search, Traditional MCTS, RF-MCTS, LogprobEvaluator, NormLenEvaluator, Reward evaluators, JudgeEvaluator, REPLEvaluator, TournamentEvaluator, RMaxTSEvaluator \\
\bottomrule
\end{tabular}
\caption{
Coverage of \textsc{TreeThink} components in our evaluation. Main experiments use
representative configurations, while the remaining components are verified
through integration smoke tests.
}
\label{tab:component_coverage}
\end{table}



In our formal proof search experiments, we select \texttt{NormLenEvaluator} as our evaluator due to its lightweight and deterministic features. For informal language test, we use \texttt{JudgeEvaluator} to demonstrate our framework's ability to integrate reward models into the search process.





\section{Experiment Variables for Reproducibility Purposes}
\label{sec:reproducibility}

\begin{table}[h]
    \centering
    \small
    \renewcommand{\arraystretch}{1.2}
    \begin{tabular}{@{}p{0.4\linewidth}p{0.5\linewidth}@{}}
        \toprule
        Parameter & Value \\
        \midrule
        TreeThink commit hash & 4887eb86a15f2b2bd5c4
        cf2fa0bdd34502eac357 \\
        TreeThink PyPI package version & v0.1.0 \\
        vLLM version & v0.21.0 \\
        Lean/Rocq/Isabelle versions & v4.21.0 / 9.0 / 2025-2 \\
        REPL server versions & v1.0.1 / v0.1.0 / v0.3.5 \\
        Exact HuggingFace model IDs & deepseek-ai/DeepSeek-Prover-V2-7B, Qwen/Qwen2.5-Coder-14B-Instruct, Qwen/Qwen3.5-9B \\
        Exact HuggingFace dataset IDs & HaimingW/miniF2F-lean4 / LLM4Rocq/miniF2F-rocq / erer-can/minif2f-isabelle \\
        Sampling parameters & temperature: 0.3, top\_k: -1, top\_p: 0.95 \\
        Timeout behavior & Max. 480s for a single proof search. Max. 120s for REPL . \\
        Hardware (CPU/RAM/GPU) & AMD EPYC 9224 24-Core Processor / 32 GB RAM / Nvidia A40 \\
        Is averaged over multiple runs? & All experiments are single runs, except for \S 4.2, which is averaged over 4 runs. \\
        \bottomrule
    \end{tabular}
    \caption{Experiment details: version numbers, HuggingFace model/dataset IDs, sampling parameters, timeout behavior, hardware and whether the experiments are averaged over multiple runs or not.}
    \label{tab:exp_config}
\end{table}

The list of experiment variables can be found in Table \ref{tab:exp_config}. For the datasets we use, we construct columns containing both the header and the problem statement. To incentivize proof generation by the model, we modify the target language syntax: replacing \texttt{sorry} with \texttt{by} in Lean 4 and with \texttt{proof} in Isabelle, while appending the \texttt{Proof.} keyword to Rocq statements.




\end{document}